\documentclass[11pt, a4paper, logo, address]{thuair}

\usepackage[numbers, compress]{natbib}

\usepackage{nicefrac}       % compact symbols for 1/2, etc.
\usepackage{mwe}
\usepackage{array}
\usepackage{float}
\usepackage{makecell}
\usepackage{subcaption}
\usepackage{xspace}
\usepackage{minted}
\usepackage{wrapfig}
\usepackage{tcolorbox}
\usepackage{fontawesome5}
\tcbuselibrary{breakable, skins}

\title{{\sys}: Guiding GUI Agents with Coding Agents via Large-Scale Environmental Synthesis}

\author[1]{Guohong Liu}
\author[2]{Jialei Ye}
\author[3]{Pengzhi Gao}
\author[3]{Wei Liu}
\author[3]{Luan Jian}
\author[1]{Yunxin Liu}
\author[1]{Yuanchun Li\textsuperscript{$\dagger$}}
\affil[1]{Institute for AI Industry Research (AIR), Tsinghua University}
\affil[2]{University of Electronic Science and Technology of China}
\affil[3]{MiLM Plus, Xiaomi Inc.}

\correspondingauthor{liyuanchun@air.tsinghua.edu.cn}

\newcommand{\sys}{ScaleWoB\xspace}
\newcommand{\eg}{e.g.\xspace}
\newcommand{\etc}{etc.\xspace}
\definecolor{myframe}{HTML}{118C7E}
\newcolumntype{P}[1]{>{\raggedright\arraybackslash}p{#1}}
\newcolumntype{C}[1]{>{\centering\arraybackslash}p{#1}}
\newcolumntype{R}[1]{>{\raggedleft\arraybackslash}p{#1}}

\begin{document}

% thuair.cls typesets the abstract inside \maketitle, so it must be
% declared before \maketitle (unlike the NeurIPS order).
\begin{abstract}
GUI agents powered by large language models are advancing rapidly, 
creating urgent needs for evaluation and training based on realistic environments.
However, directly doing so in real-world environments introduces 
some challenges that cannot be overlooked. Real-world environments are complex and 
uncontrollable, making it difficult to construct verifiable rewards and to save or reset states.
Existing works prioritize reproducibility but are often limited to open-source apps
or file-operation tasks for reliable reward building, leaving a persistent gap from 
real-world usage. Furthermore, relying on virtual machines or docker images demand high resource 
requirements and suffer from slow response speeds, which limit the efficiency.
We present \sys, a framework that could produce high-fidelity synthesized interactive
environments for GUI agents across platforms with verifiable rewards.
These environments behave as backend-free webpages accessible via URL, 
requiring near-zero setup and low resource cost, making the approach suitable for 
both large-scale evaluation and downstream agent training.
We support multiple GUI platforms including mobile, desktop, and automotive/in-vehicle 
interfaces based on the same pipeline, covering 100+ environments and 1000+ verifiable tasks.
Among them, 120 challenging tasks across 63 simulated mobile applications are released 
as a fully synthesized mobile GUI agent benchmark.
Experiment results on five state-of-the-art mobile GUI agents reveal substantial headroom---the
average success rate is only 27.92\%, dropping to 17.82\% on long-horizon subset---while
humans reach 92.08\%. A comparison against real-world sample tasks shows that assessments
made in our synthetic environments generalize to real apps.
\end{abstract}

\maketitle

\begin{center}
  \small \noindent{\color[HTML]{1a58ab}\faIcon{globe}} \textbf{Project}: \href{https://scalewob.github.io}{scalewob.github.io} \quad
\noindent\faIcon{github} \textbf{Code}: \href{https://github.com/ScaleWoB/ScaleWoB}{github.com/ScaleWoB/ScaleWoB}  
\end{center}

\section{Introduction}
\label{sec:intro}

GUI agents~%
\cite{autodroid,UI-TARS,Aguvis,OS-Copilot,aria-ui,android-in-the-zoo,mobile-agent-v3,agent-q,agent-s,autoglm,step-gui,mobileagentv3.5,MAI-UI}
powered by large language models~(LLMs) and vision language models~(VLMs) have progressed rapidly in recent years.
Given a task and an interactive interface, an autonomous agent must interpret the current screen, reason over task state,
execute actions step by step, and use environment feedback until completion.
To improve the performance of GUI agents, a series of related works~\cite{WebRL,MobileRL,WebAgent-R1,GiGPO} emphasize the importance of 
enabling agents to operate in interactive environments with feedback.
Besides, to measure capabilities of agents, many benchmarks~%
\cite{android-lab,android-world,os-world,CRAB,ScienceBoard,windows-agent-arena,mobile-agent-bench,MobileWorld,%
MobileEnv,weblinux,WebVoyager,pixelhelp}
have been proposed across platforms and task settings as well.
Using real-world environments for evaluation and training seems to be a very direct method,
however, it poses numerous challenges, and existing work has yet to adequately address these issues.

The first challenge is \textbf{building verifiable rewards} for tasks and environments.
In real-world environments, verifying task completion is not trivial. Many tasks involve the internal 
states of specific applications, and these states are often difficult to obtain.
To address this problem, some~\cite{WebRL,GTA1,SmartSnap,ProRe} leverage LLM-as-Judge method, using LLMs/VLMs
to evaluate the execution trajectories of agents; and some decide to use open-source 
applications with public backends~\cite{android-world,MobileWorld}, file-operation tasks~\cite{EvoCUA}, 
or execution-pattern matching that requires substantial manual efforts~\cite{android-lab,os-world,windows-agent-arena}.
These either introduces more instability and cost, or constraints narrow task coverage and 
create a gap between benchmark tasks and real-world app usage scenarios.
The second challenge is \textbf{state management and resuming}.
The evaluation and training of GUI agents rely on the stability of tasks and environments, 
requiring both reproducibility and traceability. In real-world environments, state resetting is 
nearly impossible to achieve, as many system components and contents change over time, such as 
live information feeds and irreversible operations (e.g., payments, deletions).
As a result, many have to depend on emulators, virtual devices, or Dockerized systems~\cite{android-world,os-world,windows-agent-arena,WebArena}
for state loading, resetting and recoveries.
While effective, this setup increases system complexity and runtime overhead, slowing large-scale
evaluation and downstream tasks such as online RL trainings,
and still cannot cover all real-world cases.
We address these limitations with {\sys}, and reframe environment construction itself as the bottleneck to solve.

\begin{figure}[H]
    \centering
    \includegraphics[width=0.8\textwidth]{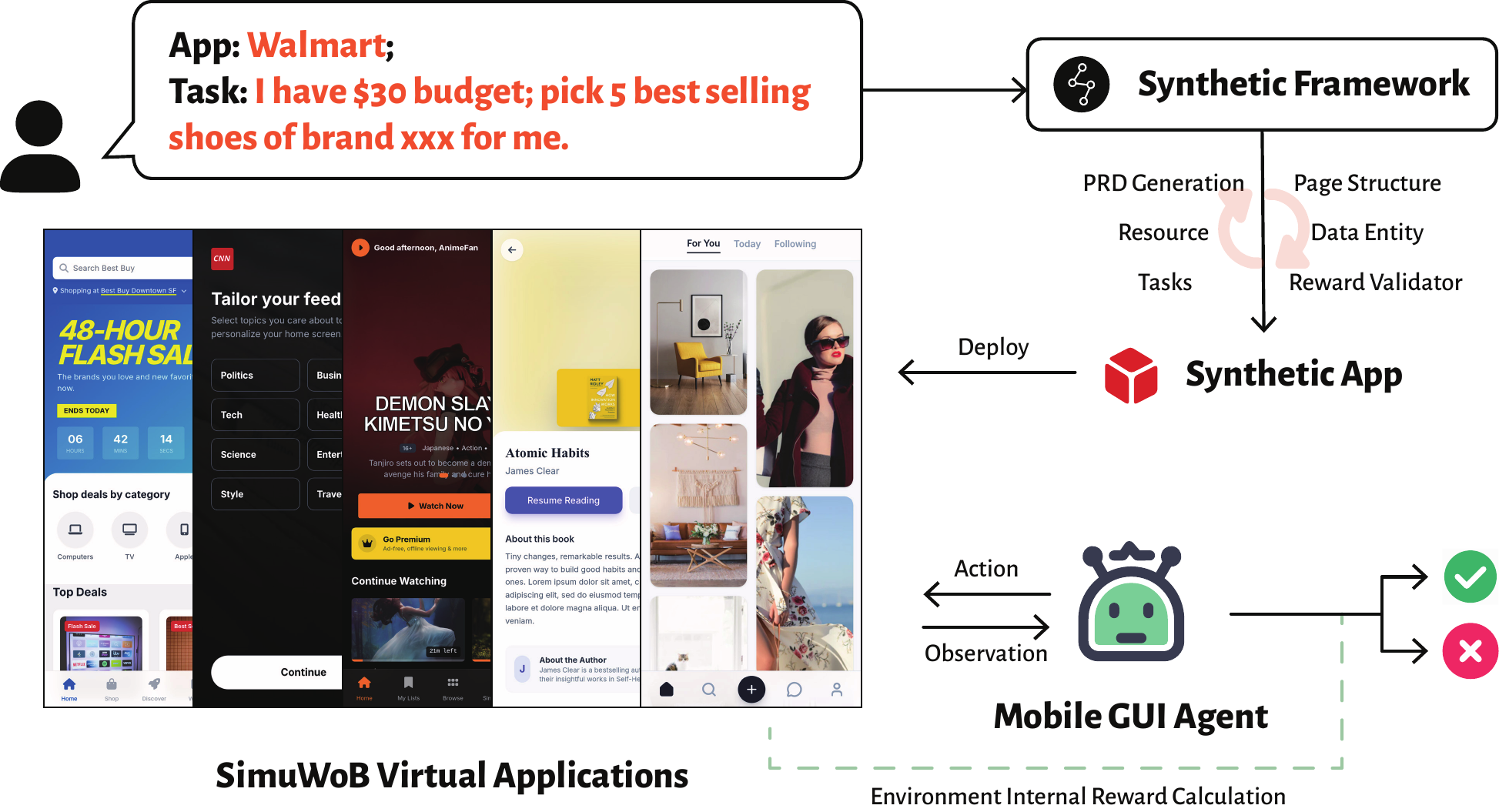}
    \caption{An overview of {\sys}. After collecting real-world tasks along with related applications,
our framework iteratively builds and refines the environment until the task and reward validator
are ready. GUI agents can then directly interact with the environments, perform tasks and receive rewards.}
    \label{fig:overview}
\end{figure}

{\sys}~(Scale World-of-Bits) is a framework that puts \emph{coding agents} to work building the worlds in which \emph{GUI agents} are measured.
Leveraging the code-generation abilities of modern LLMs%
~\cite{llm-codegen-survey,code-foundation-model-agents,software-dev-life-cycle,challenges-paths-ai-software},
it takes a natural-language task description and optional screenshots of the target workflow,
then iteratively generates and refines a backend-free webpage that simulates a real-world application
with aligned interaction logic over mock data, as illustrated in Figure~\ref{fig:overview}.
During generation it also validates task solvability and emits an executable, reliable reward function for each task.
This design directly improves \textbf{environment coverage}, \textbf{reward reliability}, and \textbf{evaluation speed},
circumventing the challenge of constructing rewards inherent in real-world apps,
enabling more faithful benchmarking at a much lower operational cost.
Besides, the framework is \textbf{platform-agnostic by construction}:
the same synthesis pipeline produces interfaces for mobile, desktop, and automotive/in-vehicle (IVI) interaction styles,
and live demos across these platforms are available on our project website.
We have created over 100 environments and over 1000 verifiable tasks across different platforms.
Among these tasks, we select 120 tasks across 63 simulated mobile app environments, spanning 20 of 33 Google
 Play Store app categories, constructing a fully synthesized and challenging mobile GUI agent benchmark.
All tasks and scenarios are collected based on a user study covering 86 participants across 77 industries.
Every environment is served via a URL, which enables lightweight deployment and efficient evaluation in parallel
with near-zero setup overhead. To probe different capability dimensions, we organize tasks into three categories:
\emph{simple}, \emph{long-horizon}, and \emph{math-related}.
Half of the tasks require more than 20 interaction steps, and the most challenging ones require over 50 steps to finish.
We benchmark five recent state-of-the-art mobile GUI agents%
~\cite{UI-TARS,autoglm,step-gui,Seed1.8,gemini}.
Results show substantial headroom for current agents: the average success rate is 27.92\% over all tasks and only 17.82\% on the long-horizon subset, while humans reach 92.08\%.
Besides, a comparison against 20 sample tasks selected from the real world demonstrates that agent assessments based on our synthetic environments generalize well.

Our contributions are as follows:
\begin{enumerate}
    \item We introduce \sys, a scalable, platform-agnostic framework in which coding agents synthesize interactive, verifiable GUI environments and tasks directly from natural-language descriptions, for efficient agent evaluation and training.
    \item We construct a fully synthesized mobile GUI agent benchmark by selecting 63 applications and 120 tasks spanning multiple languages, task formats, and difficulty levels, each shipping with an automatically constructed reward validator.
    \item We evaluate five state-of-the-art mobile GUI agents and reveal large performance gaps on complex tasks, especially long-horizon ones, with detailed analysis of failure modes and implications for future agent development.
\end{enumerate}

\section{Related Work}
\label{sec:related_work}
GUI agents operate digital interfaces in the same loop as humans:
observe the current UI state, reason about intent and progress, and execute actions across desktop, web, and mobile platforms.
With the rise of LLMs and VLMs~\cite{llm-survey,survey-of-llms}, recent work has rapidly expanded agent capabilities~%
\cite{autodroid,UI-TARS,Aguvis,aria-ui,mobile-agent-v3,agent-q,agent-s,autoglm,step-gui,mobileagentv3.5,MAI-UI,CogAgent}.
Early systems mainly relied on text-only interface representations~\cite{autodroid,mind2web,AutoWebGLM,WebAgent},
while newer agents increasingly consume screenshots and other visual signals~%
\cite{UI-TARS,aria-ui,mobile-agent-v3,autoglm,step-gui,OS-ATLAS,CogAgent,OmniParser,UGround}.
From a systems perspective, existing approaches broadly include modular pipelines and more end-to-end policies that directly map multimodal observations to actions.
Recent surveys~\cite{cua-survey,osagents} further highlight that robust computer-use agents require stronger long-horizon planning, memory, grounding reliability, and stable execution under noisy interface states.

Benchmarks for GUI agents can be grouped into {static datasets} and {interactive environments}.
Static datasets~\cite{pixelhelp,SeeClick,OS-ATLAS,ScreenSpot-pro,mind2web,aitw,gaia,seq2act,motif,meta-gui,OmniACT,AndroidControl}
are valuable for scalable offline evaluation of grounding, instruction following, and action prediction.
However, they generally do not capture closed-loop interaction dynamics (\eg, recovery from mistakes, delayed feedback, or stateful multi-step dependencies).
Interactive benchmarks~\cite{android-lab,android-world,os-world,mobile-agent-bench,MobileWorld,WebArena,agentbench,miniwob++,webshop,visual-WebArena,WorkArena,wikihow,Android-Agent-Arena}
provide executable environments and therefore better measure end-to-end success.
These have substantially advanced reproducible evaluation, but also expose a practical trade-off:
higher realism often brings higher engineering cost in environment setup, state reset, and resource consumption.
Many benchmarks depend on containers, emulators, or VM snapshots to preserve recoverability, which can limit evaluation speed and concurrency.
Another bottleneck is {reward construction under realistic app constraints}.
For reliable automatic scoring, many benchmarks either focus on environments with accessible internal state (\eg, open-source apps, synthetic web worlds, or file-based tasks)~\cite{android-world,MobileWorld,EvoCUA}
or use extensive manual rule engineering for execution checking~\cite{android-lab,os-world,windows-agent-arena}.
This makes it difficult to scale toward faithful simulations of real-world workflows, especially when task state is complex and not directly observable.
At the same time, a non-trivial portion of current tasks remains short-horizon or structurally simple, reducing discriminative power for stronger agents and leaving long-horizon failure modes underexplored.

A complementary line of work reduces this cost by \emph{synthesizing} environments, tasks, or interaction
data rather than authoring them entirely by hand.
One direction builds controllable but hand-engineered simulators of apps and users, such as
AppWorld~\cite{AppWorld} and {$\tau$}-bench~\cite{tau-bench}, which expose API-level simulated worlds with
state-based checks; these are reproducible, but their environments are manually implemented and confined to a
few domains.
A second direction keeps existing environments fixed and uses LLMs to synthesize tasks and trajectories within
them---by replaying web tutorials~\cite{AgentTrek}, exploration-driven rollouts~\cite{Explorer}, reverse task
synthesis~\cite{OS-Genesis}, backward construction from documentation~\cite{learn-by-interact}, or converting
indirect knowledge into demonstrations~\cite{Synatra}---while AgentGen~\cite{AgentGen} generates environments and
tasks jointly, though for text-based planning rather than GUI interaction.
Such efforts mainly target \emph{training data}, which connects to a growing body of online RL for GUI and web
agents~\cite{DigiRL,WebRL,MobileRL,WebAgent-R1,GiGPO} that depends on interactive, resettable environments and
reliable reward signals---precisely the resources that are scarce and costly to construct for real apps.
Related generative mechanisms also appear in front-end code generation, where models render webpages from
screenshots~\cite{Design2Code}, and in using LLMs as world models that simulate interface outcomes~\cite{WebDreamer}.

Our work is positioned at this intersection, but approaches it from the supply side: rather than
hand-building one more environment suite---or synthesizing tasks and trajectories \emph{inside} environments that
still must exist---{\sys}~treats environment construction itself as a code-generation problem, and lets coding
agents synthesize backend-free, URL-accessible environments together with executable rewards.
Unlike controllable simulators that are manually engineered for narrow domains~\cite{AppWorld,tau-bench},
the resulting environments are platform-neutral webpages, so the same pipeline targets mobile, desktop, and
automotive interfaces, emphasizing \emph{faithful simulation of real-world apps} and \emph{fast benchmarking
throughput} at once.
Moreover, because each environment ships with a verifiable reward and resets at near-zero cost, it doubles as a
substrate for downstream agent training, even though this release instantiates the evaluation setting.
Compared with prior mobile and cross-platform benchmarks, this design aims to improve realism-task coverage,
reward scalability, and evaluation efficiency simultaneously, while explicitly stressing complex task categories
such as long-horizon, ambiguous, composite, and reasoning-heavy workflows.

\section{\sys}
\label{sec:method}
In this section, we introduce {\sys}, a framework in which coding agents synthesize interactive
environments and reward validators for GUI agents, together with the mobile benchmark we build with it.
In Section~\ref{subsec:env-gen}, we describe how environments are generated with our LLM-powered
pipeline and refined through an automatic feedback loop.
In Section~\ref{subsec:bench-construct}, we present the task selection and benchmark construction process
for the 120-task mobile instantiation released here.
Because every environment is a self-contained, URL-served webpage, the same pipeline extends naturally
across GUI platforms; we make this scope explicit in Section~\ref{subsec:platforms}.

\subsection{Environment Generation}
\label{subsec:env-gen}

\begin{figure}[H]
    \centering
    \includegraphics[width=0.9\textwidth]{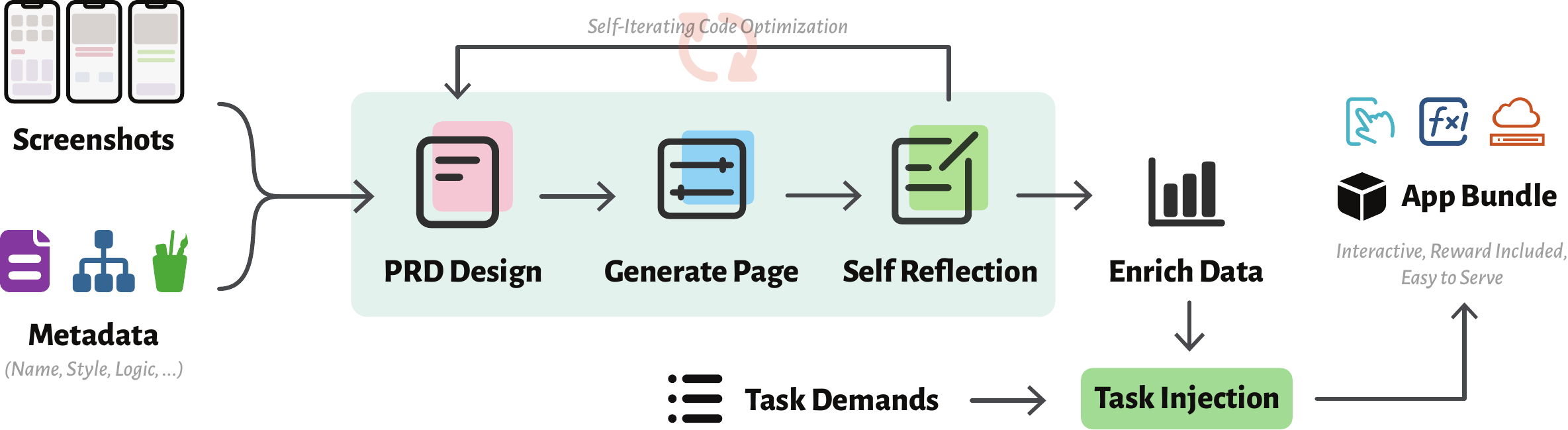}
    \caption{Two-stage environment synthesis pipeline of \sys.
Stage 1 turns app metadata and optional screenshots into a minimal working environment via iterative refinement of UI pages, data entities, and interaction logic;
Stage 2 injects task requirements, edits the environment for task executability, and synthesizes task-specific reward validators.}
    \label{fig:gen}
\end{figure}

Our goal is to build executable GUI environments that are both realistic enough to reflect real-app
interaction patterns and structured enough for large-scale, reliable rewards and evaluation.
We describe the pipeline through its mobile instantiation, though nothing below is mobile-specific
(Section~\ref{subsec:platforms}).
In practice, this requires jointly handling UI layout, interaction logic, persistent data state, 
and task-level verification, while keeping generation cost manageable across apps and tasks.
We therefore formulate environment synthesis as a 2-stage process rather than a single-pass generation,
as shown in Figure~\ref{fig:gen}.
The design principle is to first construct a realistic app simulation, and then inject benchmark tasks and validators.
This separation improves both quality control and generation efficiency: Stage 1 focuses on app fidelity and functional completeness, while Stage 2 focuses on task executability and precise automatic checking.
The following paragraphs describe these two stages and the subsequent validation-and-repair loop used to ensure final usability.

\textbf{Stage 1: Minimal working environment construction.}
We first collect application metadata from public sources, including app name, 
visual style, feature summary, and core interaction logic.
When available, additional screenshots will be provided to better align layout patterns, 
iconography, and information hierarchy with real applications.
Given these inputs, a \emph{coding agent} backed by a code-generation LLM (\eg, Gemini~\cite{gemini} or
Claude~\cite{Claude}) runs an iterative build loop.
In each iteration, the model:
\emph{(i)} drafts or updates a PRD (Product Requirements Document)
that designs the application's pages, features, design styles, \etc
\emph{(ii)} implements or revises page structure, data schema, and interaction logic,
and \emph{(iii)} performs a self-review pass over completeness and consistency.
The review output is fed back into the PRD, which drives the next implementation round.
After a predefined number of iterations, we obtain a stable minimal working environment (MWE) with executable UI logic, initial data entities, and seed mock records.
Examples of PRD and refinements are listed in Appendix~\ref{appendix:env-synthesizing}.

\textbf{Stage 2: Task injection and reward synthesis.}
Starting from the MWE, we first expand the database with richer mock content (texts, images, and structured records) using the same schema and style constraints as Stage 1.
We then provide task specifications that include expected execution intent and verification criteria.
A task-injection coding agent scans the codebase, patches task-relevant logic when necessary, and synthesizes executable validators for each task.
Because we control fine-grained environment state transitions, validators can check success conditions with perfect precision rather than relying only on approximate pattern matching.
Besides, the two-stage design also improves environment quality.
By constructing app logic before task-specific editing, the generated environment is less likely to overfit onto a single target trajectory.
In other words, the environment remains broadly usable beyond one scripted path.
To further reduce task-path overfitting, we co-generate related tasks under the same app context instead of injecting isolated tasks one by one.

\begin{figure}[H]
    \centering
    \includegraphics[width=0.8\textwidth]{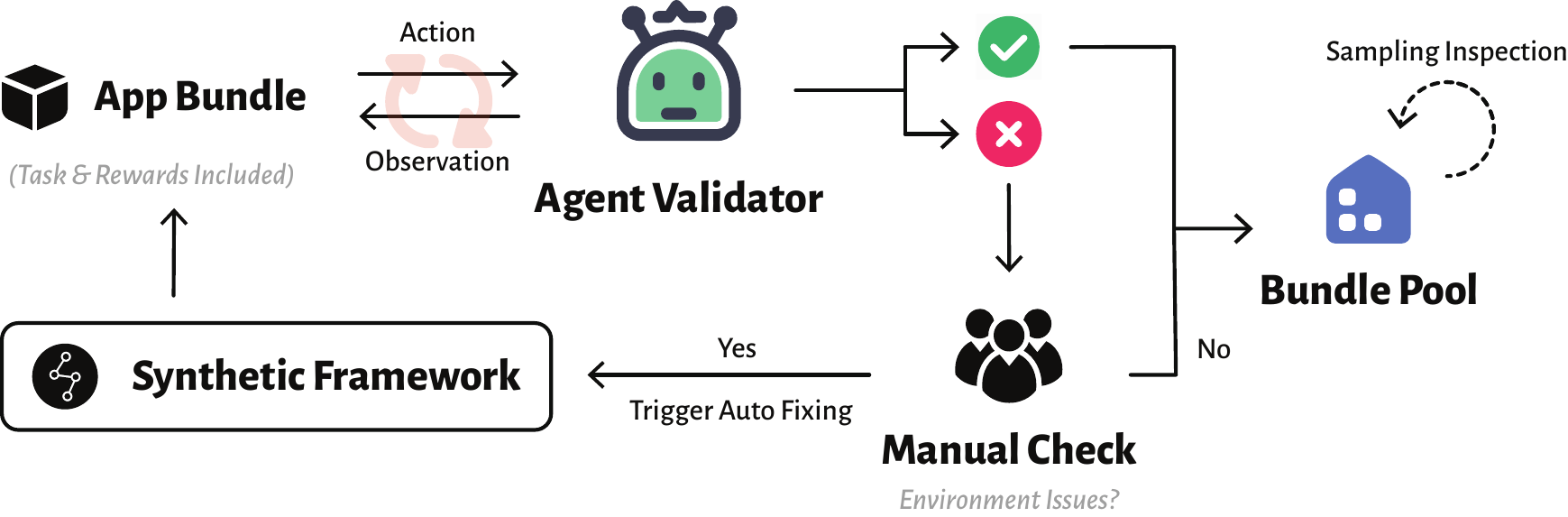}
    \caption{Automatic issue inspection and correction workflow.
A validation agent executes each bundle's tasks in the synthesized environment, and failed trajectories are escalated to human experts for root-cause diagnosis.
Environment-side issues return to the repair loop, while passing bundles enter the final pool and undergo random quality audits.}
    \label{fig:auto-validation}
\end{figure}

\begin{figure}[H]
    \centering
    \includegraphics[width=0.9\textwidth]{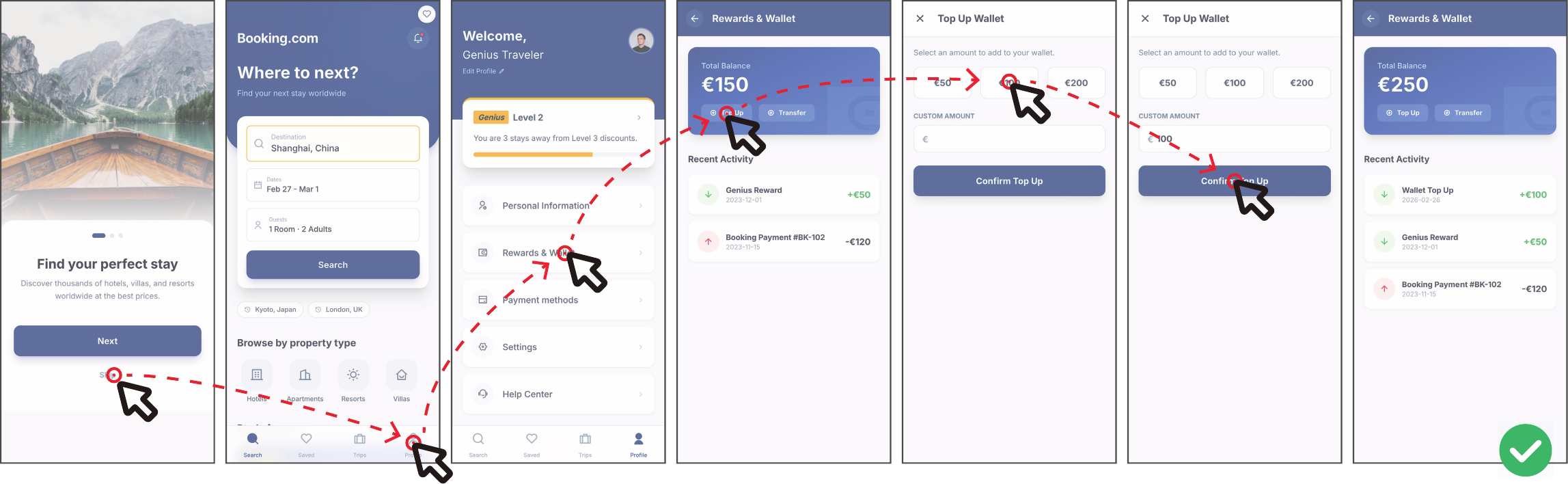}
    \caption{Example task in \sys.
The agent is asked to top up the wallet by 100 euros in a simulated hotel-booking app.
The generated environment closely matches real-app appearance and interaction behavior, and persists state via local storage for consistent multi-step execution.}
    \label{fig:task-example}
\end{figure}

Considering that large language models, when generating complex environments, are constrained by their 
inherent capabilities and potential hallucinations, the resulting environments cannot guarantee 100\% usability.
For instance, they may contain flawed interaction logic or UI design issues that prevent certain tasks from being completed. 
To address this, we designed a human-in-the-loop issue detection and repair mechanism that remains scalable and 
applicable even in large-scale generation scenarios. The main workflow of this is shown in Figure~\ref{fig:auto-validation}.
For each generated app bundle, we run a multi-step verification procedure.
For every task in the bundle, a validation agent executes the task interactively, and the synthesized validator determines success or failure.
Successful tasks are provisionally accepted.
Failed trajectories, together with environment artifacts, are sent to human experts for triage.
Experts determine whether failure is caused by agent behavior or by environment/task defects.
If environment-side defects are identified, experts provide targeted feedback, and the bundle returns to the generation pipeline for repair and re-validation.
Only bundles that pass this loop are moved into the candidate benchmark pool.
We then perform additional manual quality control via random sampling to inspect usability, logical consistency, and task reasonableness.
For the final benchmark release, \textbf{all environments and tasks are manually verified} to ensure rigorous quality and reliable experimental conclusions.
Figure~\ref{fig:task-example} shows an example generated environment and task.

\subsection{\sys~Benchmark}
\label{subsec:bench-construct}

\subsubsection{Tasks \& Environments}
\sys~is constructed with a user-need-driven pipeline.
To anchor the benchmark in real-world mobile use cases, we conducted a user study and collected open-ended task requests describing participants' daily demands.
After filtering malformed records, the study pool contains 260 valid requests from 86 participants across 77 industries.
Each record includes a natural-language user command, background pain points, and an annotator judgment of whether the request is currently feasible for a mobile agent.
We then transform all raw requirement records into benchmark tasks through four steps:
\emph{(i)}~We normalize requests by merging semantically equivalent intents and removing near-duplicates.
\emph{(ii)}~We perform feasibility screening and exclude requests that require unavailable permissions, unsupported cross-platform integrations, or non-executable conditions.
\emph{(iii)}~We operationalize retained intents into executable task specifications with explicit completion criteria.
\emph{(iv)}~We balance the benchmark across language, app domains, and difficulty dimensions.

Check Appendix~\ref{appendix:user-study} for more details of this study.
Every task is then fed into the pipeline as stated in Section~\ref{subsec:env-gen}.
In addition to the target task, we also leverage LLMs to propose other reasonable tasks within the app, 
incorporating them into the generation process to ensure the final application is more comprehensive and diverse in functionality.
% Table~\ref{tab:user-study-example} provides an example of collected real-world user requests,
% and the corresponding verified evaluation tasks included in \sys.

\begin{table}[H]
    \centering
    \caption{Examples of tasks with and without return values.
    For return-required tasks, the agent receives an additional JSON schema and must output a schema-compliant JSON object for verification.}
    \begin{tabular}{P{6em}P{16em}C{10em}}
        \toprule
        \textbf{Task Type} & \textbf{Task} & \textbf{Return Format} \\
        \midrule
        \emph{w/ returns}  & In app ..., how much have I spent on ... in 2023? & \texttt{\{"price":130.2\}} \\
        \emph{w/o returns} & In app ..., please archive all emails that meet the ... condition. & \texttt{(null)} \\
        \bottomrule
    \end{tabular}
    \label{tab:return-example}
\end{table}

Following this process, \sys~contains 120 executable tasks from 63 distinct virtual apps~(\eg, Gmail, Reddit, Spotify, Telegram), 
covering representative commercial scenarios and daily use cases.
We compiled statistics on the number of app categories covered by various mobile GUI agent benchmarks 
(based on the 33 app categories defined by the Google Play Store), where \sys~covers 20 out of the 33 app categories, 
exceeding 60\%; in contrast, other benchmarks~\cite{android-world,android-lab,MobileWorld} cover only around 30\%, 
predominantly consisting of communication and tools applications. 
This indicates that \sys~offers a more comprehensive evaluation, resulting in a much smaller gap relative to real-world usage scenarios.
The required interaction length of the final task set ranges from about 10 to over 50 steps.
Tasks in~\sys~are categorized from multiple perspectives as follows.

\emph{(1)}~\textbf{Language}. In order to evaluate cross-lingual robustness and account for language-specific UI design conventions, 
\sys~includes both Chinese and English tasks.
\emph{(2)}~\textbf{Returns}. Real-world mobile tasks often require both UI execution and structured information output.
For example, an agent may need to compute a summary value after completing several interactions.
To reflect this requirement, \sys~includes tasks with explicit return values.
For these tasks, the agent receives a JSON schema at the start and must return a schema-compliant JSON object;
both the system state and the return value provided will be considered to compute the reward.
Examples are shown in Table~\ref{tab:return-example}.
\emph{(3)}~\textbf{Task Categories}. To ensure diversity in task difficulty, 
\sys~covers simple navigation/operation tasks as well as more practical, challenging workflows.
We group tasks into 3 categories according to the main source of difficulty:
\emph{simple} (naive navigation and operations),
\emph{long-horizon} (long step chains/loops, or more information involved),
and \emph{math-related} (information aggregation and calculations).
This taxonomy enables a finer-grained analysis of agent capability under different failure modes.
Example tasks of each category and average steps are listed in Table~\ref{tab:complexity-example}.
In general, \emph{long-horizon} tasks require more steps (near 25) than the other categories.

\begin{table}[H]
    \centering
    \caption{Example tasks from different categories.}
    \begin{tabular}{P{6em}P{25em}P{3.5em}}
        \toprule
        \textbf{Task Type} & \textbf{Task} & \textbf{\# Steps} \\
        \midrule
        \emph{Simple} & Add xxx item to the cart.  & 14.96\\
        \emph{Long-Horizon} & View the first 15 images recommended in ``For You'', summarize their titles, posters, and the posters' follower counts. & 24.73\\
        \emph{Math-Related} & During my purchase history, how much did I spend on shipping in total? & 11.27\\
        \bottomrule
    \end{tabular}
    \label{tab:complexity-example}
\end{table}

\subsubsection{Evaluation}
\label{subsubsec:method-eval}

To support standardized and efficient evaluation, each environment exposes several lightweight DOM-level API functions:
\emph{(1)}~\texttt{window.getTasks}: returns task objects containing task description, 
unique id, and an optional JSON return schema.
\emph{(2)}~\texttt{window.evaluateTask}: evaluates whether the current environment state 
satisfies a target task id, with optional returned content for schema-based checking.
\emph{(3)}~\texttt{window.reset}: restores the environment to its initial state before 
action execution, preventing cross-task interference within the same application instance.
Examples of these functions are listed in Appendix~\ref{appendix:env-synthesizing}.
The same reset-and-reward contract that supports evaluation also makes each environment a resettable,
reward-bearing substrate for online RL training rather than one-shot assessment alone---the synthesis-for-training
setting discussed in Section~\ref{sec:related_work}.
% Because environments are browser-based, evaluation naturally supports headless execution and parallel workers.
Depending on hardware resources, the system can typically run 8--16 concurrent workers or more.

\subsection{Supported Platforms}
\label{subsec:platforms}

Although this release studies the mobile platform in depth, nothing in the synthesis pipeline is mobile-specific.
Environments are self-contained webpages with no backend, so a single framework can target any GUI form factor
by varying viewport, interaction conventions, and design language while reusing the same generation,
reward-synthesis, and validation machinery.
In aggregate, the pipeline has so far produced over 100 environments and over 1000 verifiable tasks across these
profiles; from this pool we curate, manually verify, and release the 120 mobile tasks across 63 apps evaluated in
this paper, while the broader set is available as live demos on our project website.
We currently expose three platform profiles:
\emph{(1)}~\textbf{Mobile}, rendered in a phone viewport with touch gestures---tap,
long press, drag, and swipe---which is the focus of the benchmark and experiments in this paper;
\emph{(2)}~\textbf{Desktop}, rendered in a standard window with pointer and keyboard interaction, suited to
productivity and web-style workflows; and
\emph{(3)}~\textbf{Automotive (IVI)}, rendered in an in-vehicle infotainment layout emphasizing large touch
targets, glanceable information, and driving-oriented tasks.
The evaluation interface is identical across profiles: an agent drives the environment through coordinate-level
actions and is scored by the same \texttt{window.getTasks}/\texttt{window.evaluateTask}/\texttt{window.reset}
contract described above.
This makes the desktop and automotive profiles drop-in extensions of the framework rather than separate systems;
we leave their large-scale curated benchmark construction and evaluation to future releases and report quantitative
results only for mobile in this work.

\section{Experiments}
\label{sec:experiments}
\subsection{Settings}
\label{subsec:settings}
We evaluate recent mobile GUI agents on \sys.
Because \sys~is currently implemented on web-based tech stacks, it provides only screenshot observations and no structured UI signals (e.g., Android Accessibility trees). Therefore, we include only agents that can solve tasks from visual input alone.
The evaluated agents are UI-TARS-1.5~\cite{UI-TARS}, doubao-seed-1.8~\cite{Seed1.8}, Gemini 3 Pro~\cite{Gemini3}, MAI-UI~\cite{MAI-UI}, and Mobile-Agent-v3.5~\cite{mobileagentv3.5} (which runs a GUI-Owl-1.5-8B backbone).
The first three are API-based models, while the latter two are open-source fine-tuned ones.
For API models, specific checkpoints are \texttt{doubao-seed-1.8-251228}, \texttt{doubao-ui-tars-250428}, and \texttt{gemini-3-pro-preview}; for local models, we deploy \texttt{GUI-Owl-1.5-8B-Instruct} and \texttt{MAI-UI-8B}.
For brevity, we refer to doubao-seed-1.8 as \texttt{seed-1.8} and Mobile-Agent-v3.5 as \texttt{GUI-Owl-1.5} in the results below.
The experiment is run with 8 parallel workers.
In preliminary trials, local fine-tuned models failed to produce schema-valid JSON outputs under prompting;
accordingly, for local models we report results only on tasks without return-value requirements.
Our primary metric is success rate (SR).
To prevent infinite loops, we cap each trajectory at 100 steps.
This cap is sufficient for task completion while keeping evaluation cost manageable.
We also evaluate human performance on these tasks as well.
All results are averaged over two runs.
Furthermore, to test whether the discriminative power of our benchmark for agents 
aligns with real tasks in actual applications, we designed 20 similar tasks from 
17 apps in a real-world environment and evaluated these agents. 
The results from the real-world environment were manually assessed to ensure accuracy.
The details of experiment setting up, real-world validation tasks, \etc are at Appendix~\ref{appendix:result}.

\subsection{Experimental Results}

\begin{figure}[H]
    \centering
    \begin{subfigure}[b]{0.45\textwidth}
        \centering
        \includegraphics[width=\linewidth]{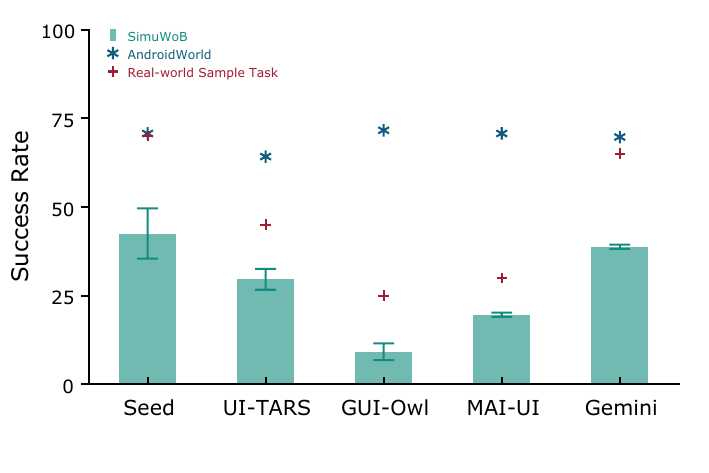}
        \caption{GUI agent SR on different benchmarks}
        \label{subfig:full-sr}
    \end{subfigure}
    \begin{subfigure}[b]{0.45\textwidth}
        \centering
        \includegraphics[width=\linewidth]{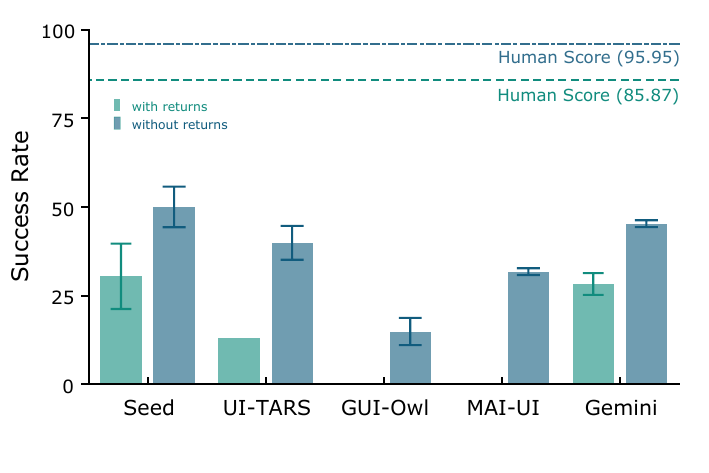}
        \caption{SR on w/ \& w/o return-value subsets}
        \label{subfig:return-value-sr}
    \end{subfigure}
    \caption{Experimental results of different agents on \sys.
    For local models, we evaluate only tasks without return-value requirements.}
    % Since Gemini 3 Pro does not have an official report, we take the score of Gemini 2.5 Pro as its lower bound.}
    \label{fig:main-result}
\end{figure}

The main results are shown in Figure~\ref{fig:main-result}. Overall, \texttt{seed-1.8} achieves 
the best SR on \sys, followed by \texttt{Gemini 3 Pro}, \texttt{UI-TARS-1.5},
\texttt{MAI-UI} and \texttt{GUI-Owl-1.5}.
For tasks without return-value requirements, \texttt{seed-1.8} reaches 50.00\%, 
\texttt{Gemini 3 Pro} reaches 45.27\% and \texttt{UI-TARS-1.5} 
reaches 39.86\%; for tasks with return-value requirements, their SR drops to 30.43\%, 28.26\% and 13.04\%, respectively;
this phenomenon may stem from the fact that tasks requiring return values are relatively more difficult.
For local models, \texttt{MAI-UI} outperforms \texttt{GUI-Owl-1.5}.
Although all agents report much higher scores on AndroidWorld (64.2\%--71.6\%), their SR on 
\sys~is substantially lower, indicating the higher challenge posed by \sys. 
In contrast, humans achieve an average score of 92.08\%, indicating a substantial gap between current mobile GUI agents and humans.

Notably, the performance trends observed on sample tasks from real apps are largely consistent with those of \sys, 
meaning the relative performance rankings among models remain the same. 
This demonstrates that evaluation results obtained in synthetic environments can generalize to real-world apps, 
thereby validating the feasibility and generalization ability of this approach.

\subsection{Analysis}

\begin{figure}[H]
    \centering
    \begin{subfigure}[b]{0.32\textwidth}
        \centering
        \includegraphics[width=\linewidth]{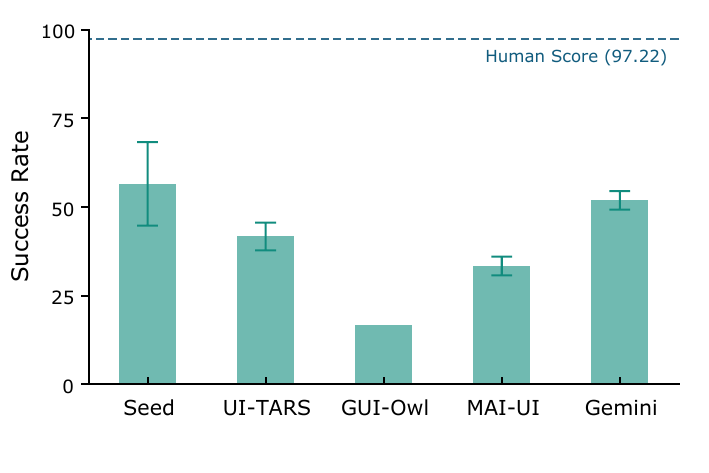}
        \caption{Simple Subset}
    \end{subfigure}
    \begin{subfigure}[b]{0.32\textwidth}
        \centering
        \includegraphics[width=\linewidth]{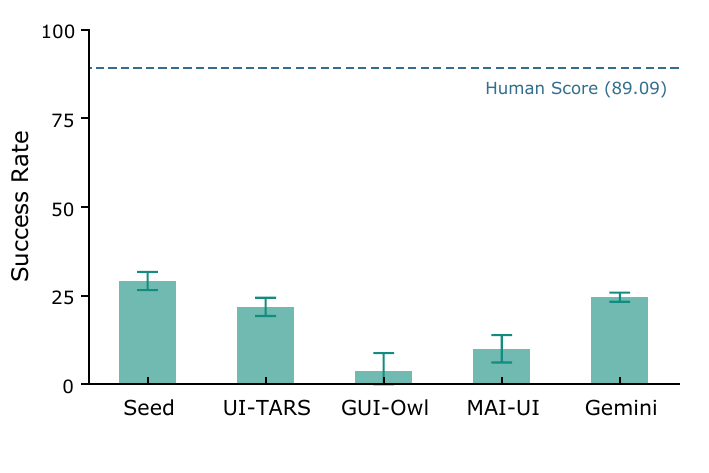}
        \caption{Long-horizon Subset}
    \end{subfigure}
    \begin{subfigure}[b]{0.32\textwidth}
        \centering
        \includegraphics[width=\linewidth]{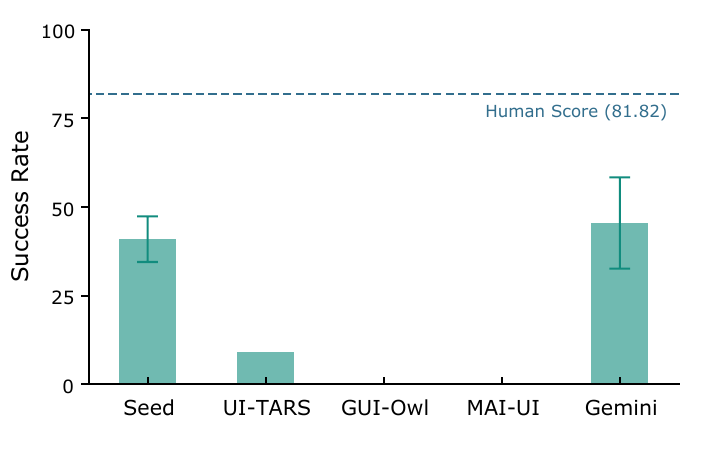}
        \caption{Math-related Subset}
    \end{subfigure}
    \caption{Success rate of different task categories across evaluated agents in \sys.}
    \label{fig:category-result}
\end{figure}

\textbf{Tasks in \sys~are challenging for mobile GUI agents.}
As shown in Figure~\ref{subfig:full-sr}, all evaluated agents achieve much higher SR
on AndroidWorld (64.2\%--71.6\%) than on \sys.
On \sys, overall SR drops to 42.50\% for the best model (\texttt{seed-1.8}) 
and to as low as 9.17\% for \texttt{GUI-Owl-1.5}; even on the easier subset without
return-value requirements, the best SR is only 50.00\%.
At the aggregate level, the average SR across all models is 27.92\%, versus
69.38\% on AndroidWorld, and this drop is consistent for all agents.
In addition, \sys~exhibits clear discriminative power: the best--worst spread in overall SR reaches
33.33 points (42.50\% vs 9.17\%), and even within stronger API models the spread remains substantial
(42.50\% vs 29.58\%). These results show that \sys~is substantially more challenging for current mobile
GUI agents and provides stronger separation across capability levels.
From Figure~\ref{subfig:return-value-sr}, we can tell that tasks with return values are relatively harder,
leading to an obvious SR drop for agents.

\textbf{Mobile GUI agents struggle on tasks requiring return values.}
Agents perform markedly better on pure operational tasks than on those requiring a structured return
(\eg, \texttt{seed-1.8}: 50.00\% vs 30.43\%; \texttt{UI-TARS-1.5}: 39.86\% vs 13.04\%).
Existing GUI agents are largely trained and optimized for UI operation, with limited capacity for information
integration and feedback.
Although some support tool calls through basic MCP~\cite{MAI-UI,autoglm} or an ``answer'' command
(\eg, \texttt{GUI-Owl-1.5} and \texttt{MAI-UI}), these typically return only natural-language descriptions;
when a schema-compliant structured output is required, such mechanisms are difficult to constrain and verify.

\begin{figure}[H]
    \centering
    \includegraphics[width=0.95\textwidth]{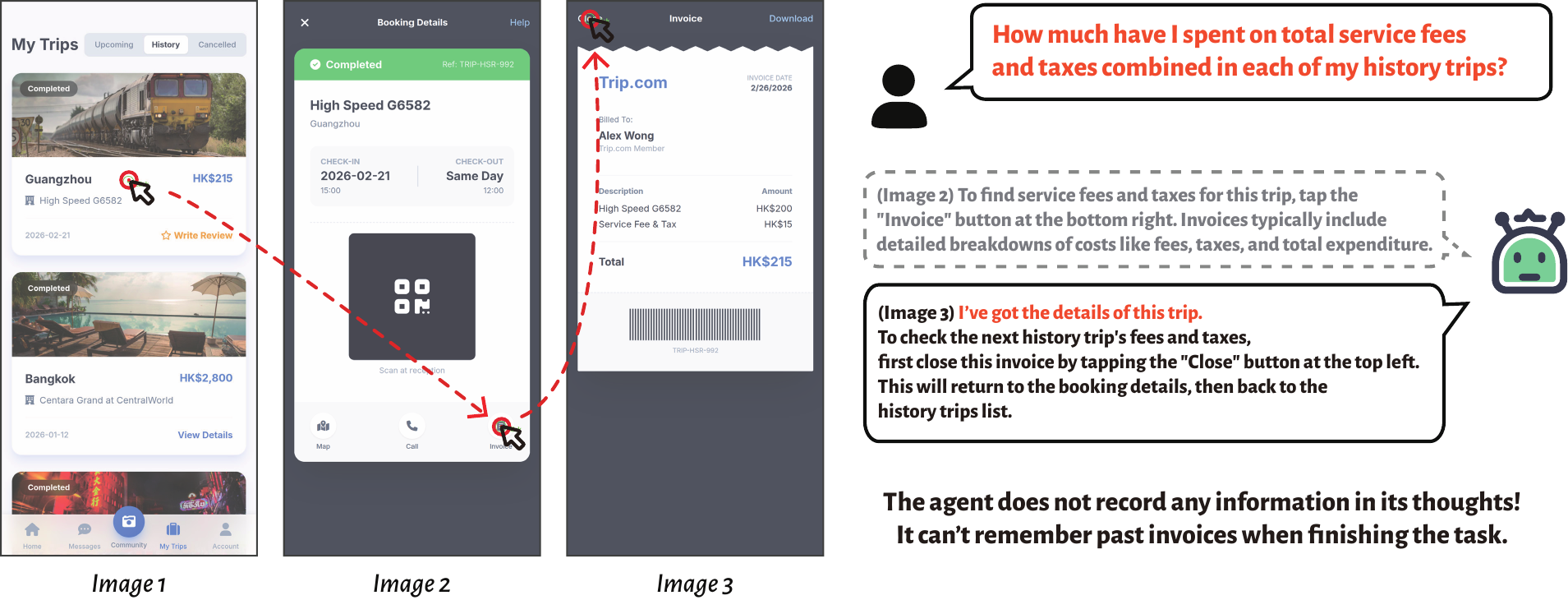}
    \caption{Case study of a \emph{long-horizon} failure: the agent executes UI operations correctly 
    but does not persist key information in context, leading to an incorrect final answer.}
    \label{fig:case1}
\end{figure}

\textbf{Mobile GUI agents fall short in \emph{long-horizon} tasks.}
Results in Figure~\ref{fig:category-result} show a clear drop on \emph{long-horizon} tasks compared with \emph{simple} tasks
for all agents (\eg, 56.48\% vs 29.09\%), with an average SR of only 17.82\%.
This gap suggests the core bottleneck is not UI operation, but state management across many steps:
agents must identify and preserve critical intermediate information while discarding low-value ones under limited context budgets.
The case in Figure~\ref{fig:case1} illustrates this failure mode. 
The task asks the agent to traverse historical trips and aggregate service fees and taxes.
The agent executes each UI step correctly (opening ``Trips'', entering history, 
opening each trip invoice), but fails to write extracted values into a persistent internal state.
It then moves to the next trip assuming previously seen details are still available, 
and eventually produces an incorrect final response.
Because most agents cannot keep all prior screenshots and observations in context at every step, 
\emph{long-horizon} performance depends on explicit memory strategies 
(what to store, when to update, and how to reuse it), which current models still handle poorly.
Similar issues also exist for \emph{math-related} tasks because agents also need to take down 
key information for further calculations.
Human SR also declines on these subsets, though far less sharply than that of agents.

\begin{figure}[H]
    \centering
    \includegraphics[width=0.95\textwidth]{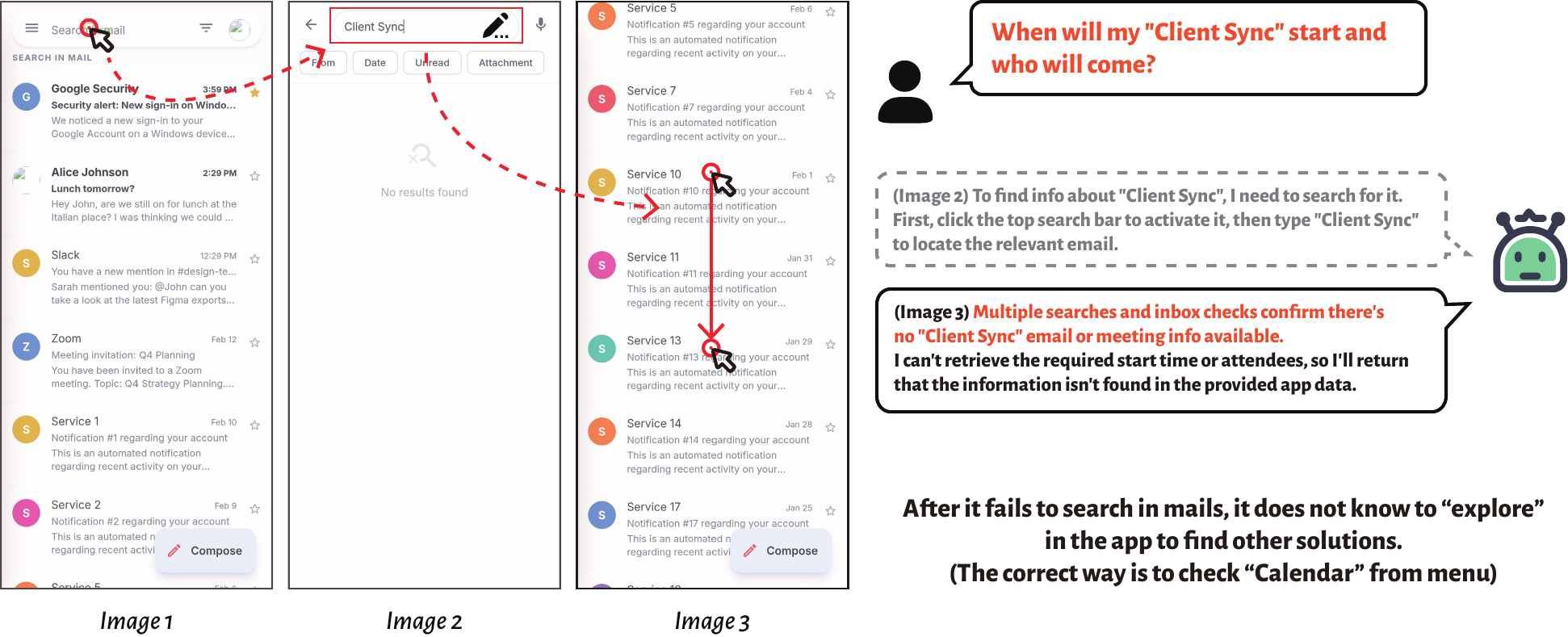}
    \caption{Case study of a \emph{vague-description} failure: the agent fails to locate 
    the task entry point due to a lack of proactive exploration capabilities following initial failures.}
    \label{fig:case2}
\end{figure}

\textbf{Tasks with vague descriptions or inconspicuous functional entry points can confuse the agent.}
Our analysis shows that agent performance degrades when instructions are underspecified and the true entry point is visually inconspicuous~(\eg, deeply nested in the UI and not exposed on the home screen).
In these settings, the key bottleneck is weak proactive exploration after an initial strategy fails.
The case in Figure~\ref{fig:case2} illustrates this behavior.
The agent is asked to find a ``Client Sync'' event in a simulated Gmail app and return its start time and participants.
Because the instruction does not specify whether the event should be searched in email or calendar views (the latter requires opening the side menu), the agent first searches in email and finds no result.
Instead of systematically switching hypotheses and exploring alternative entry points, it continues with low-yield scrolling and eventually fails.
This pattern suggests that robust exploration, trial-and-error adaptation, and online experience reuse remain underdeveloped in current agents, especially when task cues are ambiguous and interface affordances are not explicit.

\begin{wrapfigure}{r}{0.4\textwidth}
    \centering
    \includegraphics[width=\linewidth]{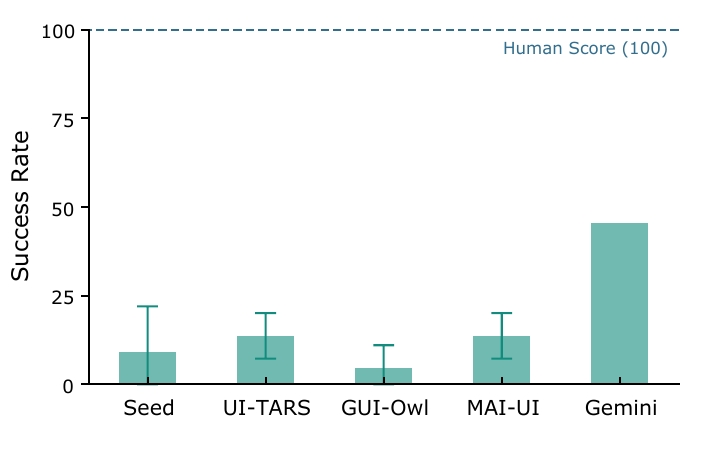}
    \caption{{Fine-grained control} results.}
    \label{fig:case3}
\end{wrapfigure}

\textbf{Agents perform poorly on tasks that require fine-grained control.}
Fine-grained control is a common requirement in real-world mobile tasks, including dragging a slider to a target position, setting date/time values with pickers, confirming payment via drag gestures, and invoking context menus through long presses.
Compared with standard click-and-type tasks, these operations impose much stricter spatial and temporal constraints: small coordinate errors or incorrect interaction timing are often enough to cause failure.
We curated 11 such tasks in \sys~and summarize model performance in Figure~\ref{fig:case3}.
While this subset is quite easy for humans to finish (full score on all tasks), 
the results on agents show a clear weakness: Gemini 3 Pro completes 5/11 tasks (still below 50\%), while the other agents complete only about 0.5 to 1.5 tasks on average.

We attribute this failure mode to two factors.
{(1)}~Insufficient action precision. For drag-and-drop interactions, many predicted coordinates are not accurate enough, leading to repeated unsuccessful attempts to reach the target state.
{(2)}~Limited interaction-strategy knowledge. Control semantics vary across interfaces: some scrollbars respond to taps, others require dragging, and some hidden actions are exposed only by long press.
Without robust UI priors or effective online trial-and-error adaptation, agents struggle to discover and execute the correct control policy.

\section{Limitations}
\label{sec:limitations}
While {\sys} synthesizes interactive environments and automatically constructs their rewards, and our
released benchmark already provides informative signals for agent evaluation, several limitations remain.
{(1)}~Observation space. Since our environments are web-based, they currently expose only
visual observations and the interaction affordances available on the page, rather than richer structured
signals (\eg, accessibility trees) exposed by some native platforms.
{(2)}~Cross-app task support. At present, \sys~focuses on single-app workflows and does not yet
cover tasks that require coordinated operations across multiple applications.
{(3)}~Platform coverage of the released benchmark. While the synthesis framework is platform-agnostic
and we provide desktop and automotive demonstrations, the quantitative benchmark in this release is
instantiated on mobile; building out and evaluating the desktop and automotive task suites at scale
remains future work.
{(4)}~Training validation. While each environment exposes a resettable, reward-bearing interface
suitable for online RL, this release empirically validates only the evaluation setting; demonstrating
downstream training gains is left to future work.
However, these constraints should be interpreted as scope boundaries rather than invalidating factors.
They do not undermine the framework's practical value, and they do not affect the main empirical
conclusion of this work: current GUI agents still show substantial gaps on realistic,
long-horizon tasks.
More broadly, the field continues to face objective bottlenecks in benchmark construction,
including environment selection, reliable reward design, and evaluation efficiency.
Our synthesis pipeline provides a technically viable foundation to address these,
and we leave the remaining items to future work.

\section{Conclusion}
\label{sec:conclusion}
In this paper, we present \sys, a framework in which coding agents synthesize the interactive
environments used to evaluate and train GUI agents.
From natural-language task descriptions, \sys~automatically constructs backend-free, URL-served
environments and task-specific reward validators, and because those environments are plain webpages,
the same pipeline spans mobile, desktop, and automotive interfaces.
In this release we instantiate the framework on mobile, yielding 63 simulated applications and 120 tasks
derived from real-world usage scenarios across 3 task categories, which we show generalize well to real apps.
Comprehensive experiments on recent mobile GUI agents show substantial headroom for current systems,
especially on long-horizon tasks, and our analysis surfaces concrete weaknesses in proactive exploration
and fine-grained action control.
We hope \sys~can serve as a scalable, practical foundation for measuring progress and guiding the
next generation of GUI agent research across platforms.

{\small\bibliographystyle{unsrtnat}
\bibliography{ref}}

\appendix
\section{\sys~Environment Synthesizing}
\label{appendix:env-synthesizing}

Following the pipeline of Figure~\ref{fig:gen}, we first had the model draft a detailed PRD document 
based on the given metadata, then asked it to write code based on the document. 
Here follows an example of PRD document when generating simulated application iQIYI, which consists of
many parts that defines the workflow, UI transition logic, data entity, styles, \etc

\begin{tcolorbox}[title=Product Overview,colframe=myframe,breakable,fontupper=\footnotesize]
    \textbf{Product Positioning}
    \vspace{0.5em} \\
    iQIYI is a leading high-quality video entertainment service platform. 
    Its core positioning is to provide long-form video streaming services (movies, TV series, variety shows, anime), 
    while gradually integrating short-form video feeds, pan-entertainment communities, and IP-derived consumption 
    (literature, e-commerce, ticketing).
    \vspace{0.5em}

    \textbf{Core Value}
    \begin{itemize}[leftmargin=*, labelsep=1em, itemindent=0pt]
        \item \emph{Enjoy Quality}: Emphasizes high-definition picture quality, Dolby audio, and exclusive original content.
        \item \emph{Immersive Experience}: Enhances viewing focus with dark mode and distraction-free playback design.
        \item \emph{VIP Privileges}: Establishes a comprehensive membership tier and benefits system, providing differentiated services.
    \end{itemize}
\end{tcolorbox}
\begin{tcolorbox}[title=Design Philosophy \& Style,colframe=myframe,breakable,fontupper=\footnotesize]
    \textbf{Overall Style}
    \begin{itemize}[leftmargin=*, labelsep=1em, itemindent=0pt]
        \item \textit{Browsing Mode (Light)}: On pages such as ``Home'', ``Discovery'', and ``Profile'', a large area of white and 
        light gray backgrounds is used. The design pursues breathability, reduces the use of dividing lines, and uses 
        card shadows and spacing to distinguish content, reducing visual pressure caused by information density.
        \item \textit{Immersive Mode (Dark/Gold)}: On the ``Playback'' page and ``VIP Member'' page, 
        the interface switches to dark or black-gold color schemes. The dark background highlights the video, 
        creating a cinema-like atmosphere; the black-gold tones reinforce the sense of premium 
        status and privileges for VIP users.
    \end{itemize}

    \textbf{Visual Metaphors}
    \begin{itemize}[leftmargin=*, labelsep=1em, itemindent=0pt]
        \item \textit{Cards}: Simulate physical posters, carrying content covers.
        \item \textit{Flow}: Waterfall flow layout, metaphorically representing an endless supply of content.
    \end{itemize}
\end{tcolorbox}
\begin{tcolorbox}[title=Main Interface Architecture,colframe=myframe,breakable,fontupper=\footnotesize]
    \textbf{Home - Content Distribution Center}
    \begin{itemize}[leftmargin=*, labelsep=1em, itemindent=0pt]
        \item \textit{Layout Logic}: Top search + navigation → Header banner carousel → Double-column waterfall recommendation feed.
        \item \textit{Key Elements}: 
            \begin{itemize}[leftmargin=*, labelsep=1em, itemindent=0pt]
                \item Top Function Area: Contains search box (displaying trending search terms), history entry, and upload/publish entry.
                \item Navigation Bar: Horizontally scrollable tabs (Recommendations, TV Series, Movies, Anime, \etc).
                \item Content Cards: Uses 16:9 cover images, with the drama title, subtitle (e.g., "All 36 Episodes"), and badges (e.g., "Exclusive," "Free for a Limited Time") displayed at the bottom.
            \end{itemize}
        \item \textit{Design Purpose}: Maximize content exposure and guide user clicks through algorithmic recommendations.
    \end{itemize}
    \textbf{Free / Suike (Short Video) - Immersive Short Videos} (...)
    \vspace{0.5em} \\
    \textbf{VIP (Membership) - Monetization Center} (...)
    \vspace{0.5em} \\
    \textbf{Discovery - Community \& Topics} (...)
    % \vspace{0.5em} \\
    % \textbf{Profile - Personal Service Hub} (...)
\end{tcolorbox}
\begin{tcolorbox}[title=Core User Flows,colframe=myframe,breakable,fontupper=\footnotesize]
    \textbf{Long-Form Video Watching Flow}
    \begin{enumerate}[leftmargin=*, labelsep=1em, itemindent=0pt]
        \item \textit{Trigger}: User clicks on a drama/movie cover on the [Home] page.
        \item \textit{Transition}: The page slides right to enter the [Playback Details Page].
        \item \textit{Loading}:
            \begin{itemize}[leftmargin=*, labelsep=1em, itemindent=0pt]
                \item Non-member: Pre-roll ad plays before the video (showing countdown and ``Skip ad with membership'' button).
                \item Member: Enters the main video directly, with a prompt saying ``Exclusive VIP, ads have been skipped for you''.
            \end{itemize}
        \item \textit{Actions}: Single-tap on the video area to evoke the control layer; use gravity sensor or tap the button to switch to [full-screen landscape mode].
    \end{enumerate}
    \textbf{Membership Subscription Flow} (...)
    \vspace{0.5em} \\
    \textbf{Short Video Browsing Flow} (...)
\end{tcolorbox}
\begin{tcolorbox}[title=Visual Interface Guidelines,colframe=myframe,breakable,fontupper=\footnotesize]
    \textbf{Color Palette}
    \begin{itemize}[leftmargin=*, labelsep=1em, itemindent=0pt]
        \item \textit{Primary Brand Color}: iQIYI Green (\#00CC36). Represents vitality and youthfulness. Used for the logo, selected states, and primary action buttons.
        \item \textit{Secondary Colors}:
            \begin{itemize}[leftmargin=*, labelsep=1em, itemindent=0pt]
                \item VIP Gold (\#E2B376): Used for membership identifiers and paid buttons.
                \item Alert Red (\#FF3B30): Used for notification badges and limited-time flash sales.
            \end{itemize}
        \item \textit{Neutral Colors}:
        \begin{itemize}[leftmargin=*, labelsep=1em, itemindent=0pt]
            \item Backgrounds: \#FFFFFF, \#F5F5F5, \#1A1A1A
            \item Text: \#333333, \#666666, \#999999
        \end{itemize}
    \end{itemize}
    \textbf{Typography}: Chinese Fonts: ..., English Fonts: ..., Hierarchy Specifications: ...
    \vspace{0.5em} \\
    \textbf{Iconography \& Components}
    \begin{itemize}[leftmargin=*, labelsep=1em, itemindent=0pt]
        \item Icon Style: Primarily line icons, transitioning to solid style with the brand 
        color when selected. Lines are rounded and approachable.
        \item Badges: Located in the top-right or bottom-right corner of the cover, using 
        semi-transparent backgrounds or brand-colored backgrounds, with extremely small font sizes, 
        used to convey status.
        \item Buttons: Full pill-shaped buttons, emphasizing tactile comfort.
    \end{itemize}
\end{tcolorbox}
\begin{tcolorbox}[title=Design Summary,colframe=myframe,breakable,fontupper=\footnotesize]
    The design of the iQIYI App successfully balances the tension between massive content distribution and user experience.
    \begin{enumerate}[leftmargin=*, labelsep=1em, itemindent=0pt]
        \item Clear Information Hierarchy: Through the card-style design featuring ``large images with minimal text'',
        users can quickly capture the visual focus while scrolling rapidly.
        \item Contextual Design: Strictly distinguishes between the ``browsing for content'' scenario (bright, efficient) 
        and the ``watching content'' scenario (dark, immersive), aligning with user mental models.
        \item Monetization Integration: The VIP membership design is not just a functional entry point but an independent 
        visual system that effectively stimulates users' desire to pay through color psychology.
        \item Ecosystem Loop: Cleverly embeds short videos (Suike) and community (Discovery) into the bottom navigation, 
        forming a content consumption loop of ``long-form video attracts → community discussion → short-form video kills time''.
    \end{enumerate}
\end{tcolorbox}

After writing, 
it reviewed the existing codebase, proposed a series of items to be added or modified, 
updated the PRD document accordingly, and iterated repeatedly. During iterations, we collect
2 examples of self-reflection outputs of the model to reveal this progress:

\begin{itemize}[leftmargin=*, labelsep=1em, itemindent=0pt]
    \item \textit{(Iteration 1)} I have enhanced the `Store' to handle more complex user states (search history, following, daily check-in), 
    updated the `Router' to handle view cleanup (essential for timers like carousels), 
    and significantly upgraded the `HomeView', `PlayerView', `SearchView', `ShortView', and `WalletView' 
    to be fully interactive and logic-driven.
    \item \textit{(Iteration 2)}  I have added the missing `Collections', `About', and `Feedback' views, 
    updated the `Player' to handle episodes dynamically, and ensured the `Profile' and `Settings' flows 
    are fully connected. I also updated the `Store' to seed some initial collection data for a better first-run experience.
\end{itemize}

After 15 iterations, we get the minimal working example of the application, then the data will be further enriched,
and task requirements will be provided to the model to inject to the app as well.
Finally, all provided tasks will be bundled into the environment through three DOM-based API functions,
as we introduced in Section~\ref{subsubsec:method-eval}.

Listing~\ref{lst:evalreset} shows an example of \texttt{window.reset} function.
Using this function, we can reset various states in the environment, ensuring that state modifications from previous 
task executions do not interfere with subsequent task executions. Additionally, this function also resets the 
page position, ensuring the page returns to its initial state.

Listing~\ref{lst:gettasks} shows an example of \texttt{window.getTasks} function in the environment.
This function returns a series of dictionary objects, each containing information about a task supported by each environment, 
including the task ID, description, and the JSON schema of the expected return value. 
Note that the model tends to generate schemas that only contain `const' fields; 
during evaluation, we automatically filter out such fields to avoid confusing the agent being evaluated.

Listing~\ref{lst:evaltask} shows an example of \texttt{window.evaluateTask} function.
For each task, the environment has corresponding low-level code that can directly access its relevant data state; 
we use local storage to manage the underlying data and ensure data consistency. This code is very concise and intuitive,
with correct logic, so no additional manually written pattern matching or similar logic is needed to complete the evaluation.

\begin{listing}[H]
\begin{minted}[fontsize=\footnotesize,frame=single]{js}
window.reset = function () {
  localStorage.clear();
  window.location.reload();
};
\end{minted}
\caption{Example of function \texttt{window.reset}.}
\label{lst:evalreset}
\end{listing}

\begin{listing}[H]
\begin{minted}[fontsize=\footnotesize,frame=single]{js}
window.getTasks = function () {
  return [
    {
      taskId: 0,
      task: "Search for the TV series The Knockout and add it to your collections.",
      params: {
        type: "object",
        properties: {
          videoId: { type: "integer", const: 1 },
        },
        required: ["videoId"],
      },
    },
    {
      taskId: 1,
      task: `Purchase the "The Knockout Official Artbook" from the Mall. During checkout,
      create and use a new shipping address with the recipient name 'TestUser'.`,
      params: {
        type: "object",
        properties: {
          recipientName: { type: "string", const: "TestUser" },
        },
        required: ["recipientName"],
      },
    },
    {
      taskId: 2,
      task: `Customize the app appearance by changing the theme to Pink and set the
      default playback quality to '4K' in Settings.`,
      params: {
        type: "object",
        properties: {
          targetTheme: { type: "string", const: "#FF3366" },
          targetQuality: { type: "string", const: "4K" },
        },
        required: ["targetTheme", "targetQuality"],
      },
    },
    {
      taskId: 3,
      task: `Go to the Wallet's Points Mall and redeem the 1-Day VIP Pass using
      your points.`,
      params: {
        type: "object",
        properties: {
          productTitle: { type: "string", const: "1-Day VIP Pass" },
        },
        required: ["productTitle"],
      },
    },
  ];
};
\end{minted}
\caption{Example of function \texttt{window.getTasks}.}
\label{lst:gettasks}
\end{listing}

\begin{listing}[H]
\begin{minted}[fontsize=\footnotesize,frame=single]{js}
window.evaluateTask = function (params) {
  if (!params || params.taskId === undefined) return { success: false };
  const user = window.AppStore.getUser();

  switch (params.taskId) {
    case 0: {
      // Task: Collect video 1
      const isCollected = window.AppStore.isCollected(params.videoId);
      return { success: isCollected, score: isCollected ? 100 : 0 };
    }
    case 1: {
      // Task: Buy product with specific address name
      const orders = user.orders || [];
      // Find order for the specific book
      const order = orders.find((o) => o.title.includes("The Knockout Official Artbook"));
      if (!order) return { success: false, score: 0 };

      // Check if address contains the required name
      const addressMatch =
        order.address && order.address.includes(params.recipientName);
      return { success: addressMatch, score: addressMatch ? 100 : 0 };
    }
    case 2: {
      // Task: Theme Pink and Quality 4K
      const currentTheme = user.settings.themeColor;
      const currentQuality = user.settings.playback
        ? user.settings.playback.defaultQuality
        : "";
      const success =
        currentTheme === params.targetTheme &&
        currentQuality === params.targetQuality;
      return { success: success, score: success ? 100 : 0 };
    }
    case 3: {
      // Task: Redeem VIP Card
      const transactions = user.transactions || [];
      const redeemed = transactions.some((t) =>
        t.title.includes(params.productTitle),
      );
      return { success: redeemed, score: redeemed ? 100 : 0 };
    }
  }
};
\end{minted}
\caption{Example of function \texttt{window.evaluateTask}.}
\label{lst:evaltask}
\end{listing}

\section{Comprehensive Experimental Results}
\label{appendix:result}
Tables~\ref{tab:main-result} and~\ref{tab:category-scores} present the full experimental results on \sys.
During evaluation, all GUI agents were tested using the prompts, code, and other resources from their official implementations.
The experiment is run with 8 parallel workers.

\begin{table}[H]
    \centering
    \caption{Experimental results of different agents on \sys. \emph{AW} denotes AndroidWorld; since
    Gemini 3 Pro has no official AndroidWorld report, we report Gemini 2.5 Pro's score as a lower bound~(\(>\)).}
    \begin{tabular}{P{9em}C{6.5em}C{6.5em}C{6.5em}C{4.5em}}
        \toprule
        \textbf{Agent} & \textbf{SR} & \textbf{SR}~\scriptsize{(\emph{w/ returns})} & \textbf{SR}~\scriptsize{(\emph{w/o returns})} & \textbf{SR}~\scriptsize{(\emph{AW})} \\
        \midrule
        \textbf{\# of Tasks} &  120 & 46 & 74 & 116\\
        \midrule
        \texttt{seed-1.8} & \underline{42.50}~\color{gray}{(\(\pm\)7.07)} & \underline{30.43}~\color{gray}{(\(\pm\)9.22)} & \underline{50.00}~\color{gray}{(\(\pm\)5.73)} & 70.7 \\
        \texttt{Gemini 3 Pro} & 38.75~\color{gray}{(\(\pm\)0.59)} & 28.26~\color{gray}{(\(\pm\)3.07)} & 45.27~\color{gray}{(\(\pm\)0.96)} & > 69.7 \\
        \texttt{UI-TARS-1.5} & 29.58~\color{gray}{(\(\pm\)2.95)} & 13.04~\color{gray}{(\(\pm\)0)} & 39.86~\color{gray}{(\(\pm\)4.77)} & 64.2 \\
        \texttt{GUI-Owl-1.5} & 9.17~\color{gray}{(\(\pm\)2.35)} & - & 14.86~\color{gray}{(\(\pm\)3.83)} & 71.6 \\
        \texttt{MAI-UI} & 19.58~\color{gray}{(\(\pm\)0.59)} & - & 31.76~\color{gray}{(\(\pm\)0.95)} & 70.7 \\
        \midrule
        \textbf{Average} & 27.92 & 23.91 & 36.35 & 69.38 \\
        \midrule
        \textbf{Human} & 92.08~\color{gray}{(\(\pm\)0.59)} & 85.87~\color{gray}{(\(\pm\)1.54)} & 95.95~\color{gray}{(\(\pm\)1.92)} & - \\
        \bottomrule
    \end{tabular}
    \label{tab:main-result}
\end{table}

\sys~provides unified interfaces for agent-environment interactions.
The observation channel currently exposes screenshots of the environment state.
The action interface maps model outputs (action type and parameters) to executable simulator operations.
Supported operations include click, swipe, long press, type, clear text, enter, wait, \etc
This design is compatible with most existing mobile GUI agents.

\begin{table}[H]
    \centering
    \caption{Success rate of different task categories across evaluated agents. We use the first 4 letters
    to represent the corresponding category: \emph{simple}~(simp), \emph{long-horizon}~(long), and \emph{math-related}~(math).}
% Given that the base numbers are too small, we don't use percentages in this table, but instead present the results directly in the form of \texttt{x/y}.}
    \begin{tabular}{P{9em}C{7.5em}C{7.5em}C{7.5em}}
        \toprule
        \textbf{Agent} & \textbf{SR}~\scriptsize{(\emph{simp.})} & \textbf{SR}~\scriptsize{(\emph{long.})} & \textbf{SR}~\scriptsize{(\emph{math.})}  \\
        \midrule
        \textbf{\# of Tasks} & 54 & 55 & 11 \\
        \midrule
        \texttt{seed-1.8} & 56.48~\color{gray}{(\(\pm\)11.7)} & 29.09~\color{gray}{(\(\pm\)2.57)} & 40.91~\color{gray}{(\(\pm\)6.42)} \\
        \texttt{Gemini 3 Pro} & 51.85~\color{gray}{(\(\pm\)2.62)} & 24.55~\color{gray}{(\(\pm\)1.28)} & 45.45~\color{gray}{(\(\pm\)12.86)} \\
        \texttt{UI-TARS-1.5} & 41.67~\color{gray}{(\(\pm\)3.92)} & 21.82~\color{gray}{(\(\pm\)2.57)} & 9.09~\color{gray}{(\(\pm\)0)} \\
        \texttt{GUI-Owl-1.5} & 16.67~\color{gray}{(\(\pm\)0)} & 3.64~\color{gray}{(\(\pm\)5.14)} & 0.00~\color{gray}{(\(\pm\)0)} \\
        \texttt{MAI-UI} & 33.33~\color{gray}{(\(\pm\)2.62)} & 10.00~\color{gray}{(\(\pm\)3.86)} & 0.00~\color{gray}{(\(\pm\)0)} \\
        \midrule
        \textbf{Average} & 40.00 & 17.82 & 19.09 \\
        \midrule
        \textbf{Human} & 97.22~\color{gray}{(\(\pm\)3.93)} & 89.09~\color{gray}{(\(\pm\)2.57)} & 81.82~\color{gray}{(\(\pm\)0)} \\
        \bottomrule
    \end{tabular}
    \label{tab:category-scores}
\end{table}

To verify that our benchmark effectively simulates real-world app usage scenarios, 
we select a total of 20 tasks across 17 apps in a real mobile environment and evaluate 
these GUI agents on them. The test results are manually verified to ensure the accuracy of the evaluation.
These tasks still cover the categories of \textit{simple}, \textit{long-horizon}, and \textit{math-related},
as shown in Table~\ref{tab:sample-tasks}.
As demonstrated in the main text of our paper, we found that the performance trends of GUI agents on 
these sample tasks are largely consistent with those observed on our benchmark. 
This indicates that our benchmark's evaluation of agent capabilities generalizes well 
to real-world usage scenarios.

\begin{table}[H]
  \centering
  \caption{Real-World Validation Sample Tasks}
  \small
  \begin{tabular}{P{6em}P{35em}}
    \toprule
    \textbf{App} & \textbf{Task Description} \\
    \midrule
    \textbf{RedNote} & Find Lei Jun's homepage and like the second post. \\
    \textbf{Douyin} & Search for ``ice cream'' in the group-buying section and add the first item to favorites. \\
    \textbf{Weibo} & Like the latest Weibo post by the user ``Cherry''. \\
    \textbf{AMap} & Find public transportation routes from the current location to Wangfujing, and start navigation for the route with the shortest travel time. \\
    \textbf{Settings} & Without using the search function, open the ``Show pointer location'' setting in Developer Options. \\
    \textbf{Wikipedia} & Open the article explaining Europa Universalis V. \\
    \textbf{Instagram} & Scroll through Instagram Reels until you find one with over 2M likes, and like it.  \\
    \textbf{Reddit} & Search for news about the new model from Anthropic and upvote to a relevant post with more than 1k upvotes. \\
    \textbf{Google Play} & Check the most relevant ratings and reviews of app ``YouTube Kids''. \\
    \textbf{Spotify} & Search for Jay Chow and start playing his album ``Opus 12'' in random order. \\
    \textbf{RedNote} & Browse the first 10 posts recommended on the homepage, and record the author, number of likes, and number of comments for the post with the highest number of likes. \\
    \textbf{Pinduoduo} & Browse the first 10 items in the ``Billions Subsidies'' channel, and output the store name, price, and number of product reviews. \\
    \textbf{Netease Music} & Browse the songs in ``Daily Recommendations'' and output the title, artist, and album of the top three songs with the longest duration. \\
    \textbf{Hema Fresh} & In the categories (excluding "Recommended for You"), view the subcategories under each category, and output which category has the most subcategories along with the count. \\
    \textbf{Trip} & Search for flights from Beijing to Shanghai, browse the top 5 flights for each of the next 3 days, record the flight with the lowest price for each day, and output the date, departure time, airline, and ticket price. \\
    \textbf{Booking} & Search for hotels in Xiamen from May 15th to May 16th at Siming District. View the top 15 hotels and output the ones with free cancellation and over 1000 reviews. \\
    \textbf{Tiktok} & Watch 15 videos and list the creator, like count, and comment count for the top 3 most-liked videos. \\
    \textbf{Spotify} & Search for Vae and check all his releases. Output the name, number of songs and total time of each album. \\
    \textbf{eBay} & Search for iPhone and check the top 20 items. Tell me the names, shop names and maximum prices of Brand New items. \\
    \textbf{Google Play} & Check top 15 Top Charts Games, and return the app with the most raters. Tell me its name, rating score, and numbers of downloads. \\
    \bottomrule
  \end{tabular}
  \label{tab:sample-tasks}
\end{table}

\section{Broader Impacts}
\label{appendix:impact}

The broader impact of \sys~is primarily tied to its role as an evaluation and development
testbed for mobile GUI agents. On the positive side, a fast, reproducible, and controlled
benchmark can help researchers identify failures in long-horizon reasoning, memory, exploration,
and fine-grained control before such agents are deployed in real applications. Because the
environments are synthetic and backend-free, evaluation can be conducted without exposing real
user accounts, private records, payments, messages, or other sensitive app data. This may lower
the barrier to safer and more transparent analysis of mobile-agent capabilities.
At the same time, stronger GUI agents may also enable harmful automation, such as unauthorized
account operations, spam, fraud, scraping, or privacy-invasive workflows, if deployed without
appropriate safeguards. Benchmarks like \sys~could indirectly accelerate these capabilities by
making agent weaknesses easier to diagnose and improve. We therefore encourage future users of
\sys~to treat it as a controlled research benchmark rather than evidence of readiness for
unsupervised real-world deployment, and to pair progress on task success with safeguards such as
permission checks, human confirmation for sensitive actions, rate limits, and auditing.

\section{User Study}
\label{appendix:user-study}
We conducted a volunteer-based user study to collect realistic mobile-agent task requests.
Participants were asked to describe daily mobile-app tasks they would like an agent to help with,
along with brief background pain points. Participation was voluntary and uncompensated.
The study did not ask participants to provide account credentials, private messages, payment
records, or other sensitive personal data. Collected requests were manually filtered,
normalized, and transformed into synthetic benchmark tasks; no real user accounts or real
personal records are included in \sys. 
All examples reported in the paper are anonymized.
The detailed questions are displayed in Table~\ref{tab:user-study-question}.
After the information was collected, we organized the feedback provided by the volunteers and 
designed corresponding tasks based on these actual needs, as shown in Table~\ref{tab:user-study-example}.
This helps us to better understands the actual pain points of daily usage on mobile applications,
thus designing more appropriate tasks to evaluate GUI agents on a benchmark which is much closer to
the real world scenario.

\begin{table}[H]
    \centering
    \caption{Questions we asked for volunteers in our user study.}
    \small
    \begin{tabular}{P{2em}P{38em}}
        \toprule
        \textbf{No.} & \textbf{Question} \\
        \midrule
        1 & What are the main apps you typically use, and what activities do you usually perform with them? \\
        2 & While using these apps, are there any aspects you find unsatisfactory or overly cumbersome? \\
        3 & Are there any needs in your daily life or work that you wish your smartphone could support but currently cannot? \\
        4 & Are there any mobile phone features or usage methods that you know others around you use, which you feel could be helpful to you, but you don't know how to use yourself? \\
        5 & Apart from your current smartphone, do you use any other smart devices regularly (e.g., a second phone, computer, tablet, etc.)? What do you use each of these devices for? \\
        6 & On a scale of 1 to 9, how would you rate your current smartphone user experience? (1 = ``Completely unable to meet my needs, useless''; 9 = ``Meets all my needs, no need for improvement''). Please explain your reason. \\
        \bottomrule
    \end{tabular}
    \label{tab:user-study-question}
\end{table}

\begin{table}[H]
    \centering
    \small
    \caption{Example of requests from user study and actual tasks included in~\sys.}
    \begin{tabular}{P{1.5em}P{4em}P{16em}P{15em}}
        \toprule
        \textbf{ID} & \textbf{Category} & \textbf{Original Request} & \textbf{Verified Task} \\
        \midrule
        118 & Cost & Help me record all online expenses, including credit card payment records. & During my history purchases, how much did I spend on shipping in total? \\
        14 & Work & Help me check which students in the WeChat group have not submitted their assignments and remind them to complete their work. & Check the ``2025 Student'' chat, identify students who have not completed the ``Completed Assignment Relay'' and send each of them a message. \\
        \bottomrule
    \end{tabular}
    \label{tab:user-study-example}
\end{table}

\end{document}